\pdfoutput=1

\documentclass[11pt]{article}

\usepackage[final]{acl}

\usepackage{times}
\usepackage{latexsym}
\usepackage{amsmath}
\usepackage{multirow}

\usepackage[T1]{fontenc}

\usepackage[utf8]{inputenc}

\usepackage{microtype}

\usepackage{inconsolata}

\usepackage{graphicx}

\usepackage[ruled,vlined,linesnumbered]{algorithm2e}
\DontPrintSemicolon
\SetKwInput{Input}{Input}
\SetKwInput{Output}{Output}
\SetKw{To}{ to }
\SetKw{Print}{Print }
\SetKwComment{Comment}{// }{}

\usepackage[normalem]{ulem}
\useunder{\uline}{\ul}{}

\definecolor{lamdocolor}{RGB}{220,20,60}

\title{ERU-KG: Efficient Reference-aligned Unsupervised Keyphrase Generation}

\author{
    \textbf{Lam Thanh Do},
    \textbf{Aaditya Bodke},
    \textbf{Pritom Saha Akash},
    \textbf{Kevin Chen-Chuan Chang}
    \\
     Siebel School of Computing and Data Science, University of Illinois Urbana-Champaign
    \\
    \texttt{\{lamdo, abodke2, pakash2, kcchang\}@illinois.edu}
}

\begin{document}
\maketitle
\begin{abstract}
Unsupervised keyphrase prediction has gained growing interest in recent years. However, existing methods typically rely on heuristically defined importance scores, which may lead to inaccurate informativeness estimation. In addition, they lack consideration for time efficiency. To solve these problems, we propose ERU-KG, an unsupervised keyphrase generation (UKG) model that consists of an informativeness and a phraseness module. The former estimates the relevance of keyphrase candidates, while the latter generate those candidates. The informativeness module innovates by learning to \textit{model informativeness through references} (e.g., queries, citation contexts, and titles) and \textit{at the term-level}, thereby 1) capturing how the key concepts of documents are perceived in different contexts and 2) estimating informativeness of phrases more efficiently by aggregating term informativeness, removing the need for explicit modeling of the candidates. ERU-KG demonstrates its effectiveness on keyphrase generation benchmarks by outperforming unsupervised baselines and achieving on average 89\% of the performance of a supervised model for top 10 predictions. Additionally, to highlight its practical utility, we evaluate the model on text retrieval tasks and show that keyphrases generated by ERU-KG are effective when employed as query and document expansions. Furthermore, inference speed tests reveal that ERU-KG is the fastest among baselines of similar model sizes. Finally, our proposed model can switch between keyphrase generation and extraction by adjusting hyperparameters, catering to diverse application requirements. \footnote{Code and data are available at \url{https://github.com/louisdo/ERU-KG}}

\end{abstract}

\section{Introduction}

\begin{figure}[h]
    \centering
    \includegraphics[width=\columnwidth]{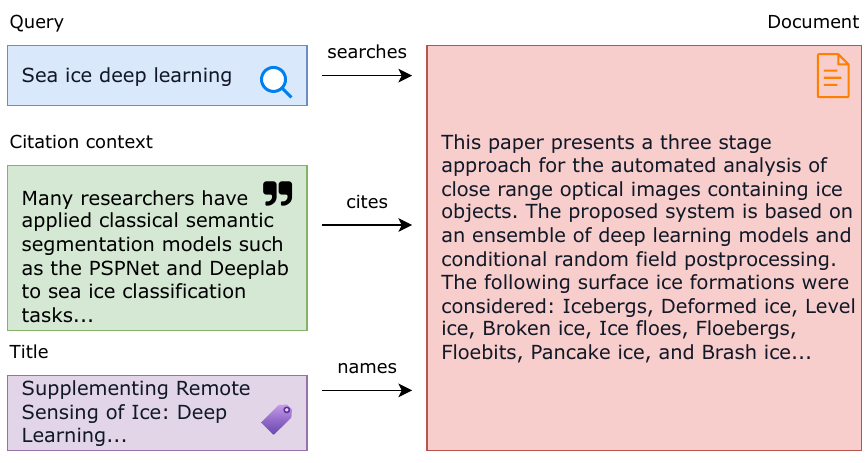}
    \caption{An example of the different type of references.}
    \label{fig:ReferencesExample}
\end{figure}

Keyphrases are short sequences of words that describe the core concepts of a document. Automatically predicting keyphrases is a crucial problem, as the outputs can be utilized in various downstream tasks, such as document retrieval \cite{zhai1997fast, gutwin1999improving, jones1999phrasier, witten2009build, fagan2017automatic, boudin-etal-2020-keyphrase} and document visualization \cite{chuang2012without}. There are two approaches for keyphrase prediction, namely \textit{keyphrase extraction} (KE) and \textit{keyphrase generation} (KG). The two approaches differ in the output space, where keyphrase generation additionally predicts absent keyphrases. Since human tend to use both present and absent keyphrases to describe documents, \textit{keyphrase generation} has received much attention in recent years.

In this work, we focus on \textit{unsupervised keyphrase generation} (UKG). In line with previous work, we target a model that receives a document as input and predicts present and absent keyphrases. The desired UKG model must learn to generate keyphrases without labeled data. Being able to build an UKG model in the unsupervised setting is highly desirable, since labeled data is often expensive and difficult to obtain. In addition, KG models are expected to be used to process large volumes of documents, as evidenced by their potential applications. For example, when utilized for document visualization or retrieval tasks, these models must efficiently handle entire corpora. Therefore, it is desirable for KG models to be \textit{time efficient}, to manage large scale data processing.

There are two challenges of building a keyphrase generation model that meet those requirements. The \textbf{first challenge} is ensuring \textit{accurate informativeness estimation}. \textit{Informativeness} refers to how well the phrase illustrates the core concepts of the text. Without labeled keyphrases, it is not straightforward to train a model that captures informativeness. Unsupervised approaches, including unsupervised keyphrase extraction (UKE) and generation, rely on heuristically designed importance scores as proxies for estimating informativeness (see Section \ref{section:related_work}). However, since these importance scores are heuristically defined, they may lead to inaccurate estimations.


The \textbf{second challenge} is \textit{efficient informativeness estimation}. Existing keyphrase generation methods typically employ a seq2seq approach that directly model the distribution of keyphrases given a document. This could make keyphrase generation slow due to the autoregressive approach taken by most models \cite{wu2022fast}. Existing UKE models, on the other hand, separate candidate phrase generation and informativeness estimation. While candidate generation is typically fast, modern UKE approaches leverage complex importance scoring function that require modeling of a document and \textit{all its candidates}, potentially slowing down the process. Specifically, embedding-based approaches \cite{bennani-smires-etal-2018-simple, sun2020sifrank} generate embeddings for the given text and all candidates, then measure informativeness via proximity in the embedding space. On the other hand, language model-based approaches \cite{ding-luo-2021-attentionrank, kong-etal-2023-promptrank} use pretrained language models (PLMs) to score each document-candidate pair individually.

Our \textbf{key idea} for addressing the \textbf{first challenge} is learning to \textit{model informativeness through references}. We propose that accurate informativeness estimation can be achieved by capturing \textit{community perception} of a document's key concepts, i.e. the central ideas as recognized by domain experts and readers. This community perception can be learned by analyzing \textit{references} - the different contexts that mention the document. We illustrate this observation in Figure \ref{fig:ReferencesExample}, where we consider three types of \textit{references}, including \textit{queries} (how the document is retrieved), \textit{citation contexts} (how the document is cited) and \textit{titles} (how the authors summarize their own work). These references provide insights into of what the community considers the key concepts of the text.

Next, our \textbf{key idea} for addressing the \textbf{second challenge} is learning to \textit{model informativeness at the term-level} rather than at the phrase-level. Estimating informativeness for each candidate phrase can be computationally expensive and slow down keyphrase generation. Instead, we propose estimating informativeness at the term-level. In particular, we leverage pairs of references and documents to train a term importance predictor, which are used to estimate informativeness of phrases by aggregating informativeness of its constituent terms, removing the need to explicitly model each candidate phrase individually.



We summarize the contributions of our paper. \textbf{Firstly}, we propose \textbf{ERU-KG}: an \textbf{E}fficient, \textbf{R}eference-aligned, \textbf{U}nsupervised \textbf{K}eyphrase \textbf{G}eneration model. ERU-KG comprises two components - an \textit{informativeness} and a \textit{phraseness module}. The former incorporates our novel key ideas to tackle the identified challenges, while the latter generates present and absent keyphrase candidates by extracting noun phrases from the given text and retrieving present keyphrases from textually-similar documents. Notably, our proposed model can switch between keyphrase generation and extraction by adjusting hyperparameters, making it suitable for different use cases. \textbf{Secondly}, we conduct \textit{groundtruth-based evaluation} and show that ERU-KG outperforms unsupervised baselines and comes very close to CopyRNN \cite{meng-etal-2017-deep}, a supervised model. \textbf{Thirdly}, to assess the utility of generated keyphrases, we carry out \textit{retrieval-based evaluation}. Our results show that keyphrases generated by ERU-KG enhance text retrieval performance when employed as query and document expansions. \textbf{Finally}, we perform inference speed test to assess the time-efficiency of ERU-KG, showing that our method is faster than existing KE and KG baselines with comparable model sizes.

\begin{figure*}[]
    \centering
    \resizebox{0.9\textwidth}{!}{

    \includegraphics[width=\textwidth]{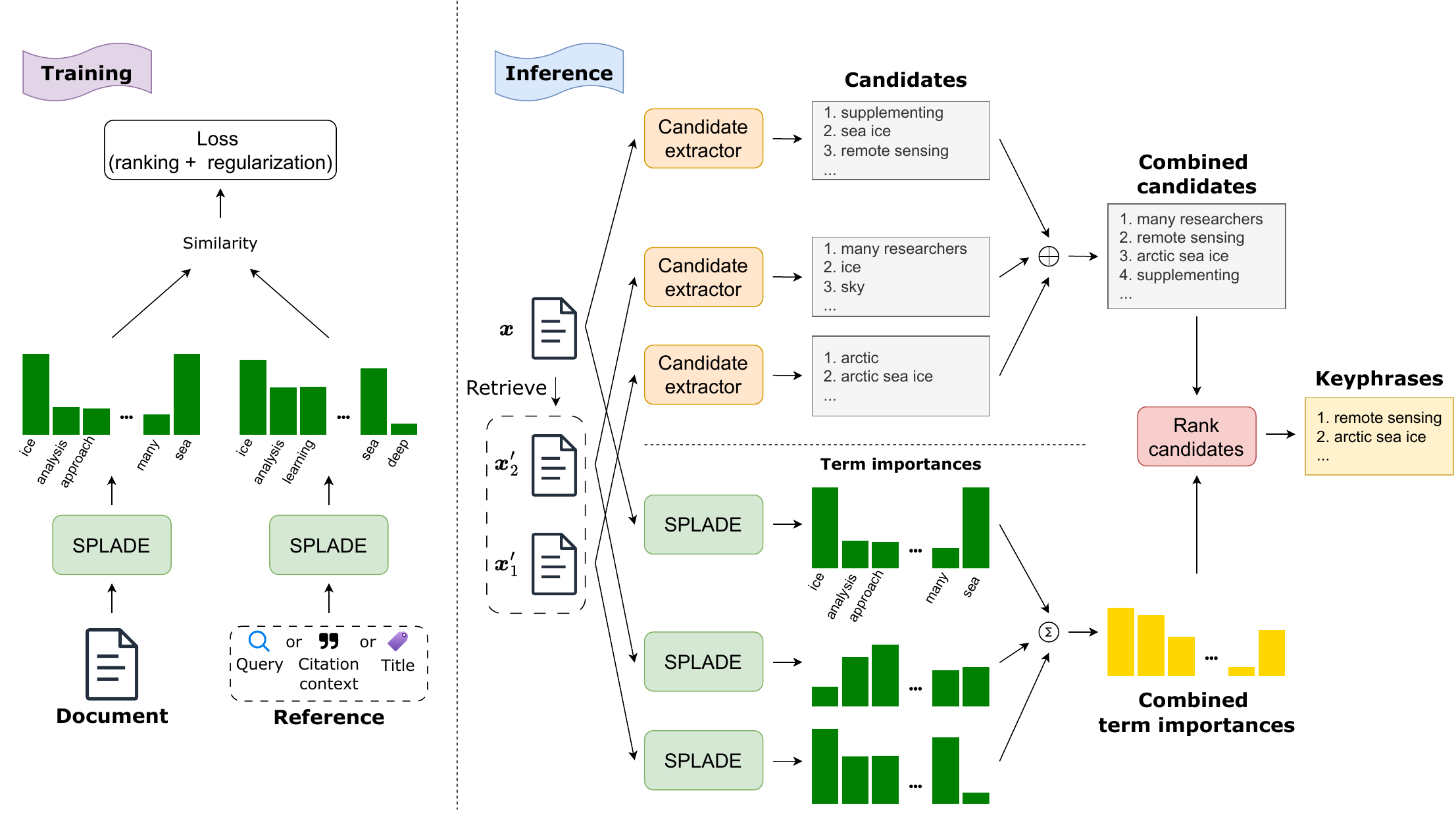}
    
    }
    \caption{Overview of \textbf{ERU-KG}. Further details of the inference process are provided in Algorithm \ref{algorithm:eru_kg_inference}}
    \label{figure:overview}
\end{figure*}

\section{Methodology}

Figure \ref{figure:overview} presents an overview of ERU-KG. Our proposed model takes as input a document $\boldsymbol{x}$ and outputs sets of present and absent keyphrases $\boldsymbol{Y}_{\boldsymbol{x}}^{\text{present}}$ and $\boldsymbol{Y}_{\boldsymbol{x}}^{\text{absent}}$, each containing $k$ keyphrases. Similar to \cite{do-etal-2023-unsupervised}, ERU-KG consists of two modules, namely \textit{informativeness} and \textit{phraseness}. The former determines the degree to which a candidate phrase represents the key concepts of the given text, while the latter is responsible for generating those candidates.


\subsection{Informativeness Module}

The informativeness module is responsible for ranking candidate phrases. As mentioned above, it incorporates our key ideas to addressing the challenges of accurate and efficient informativeness estimation: modeling informativeness through references and at the term-level. Specifically, we leverage pairs of references and documents to train a term importance predictor, which is used to estimate informativeness of candidate phrases during inference.

There exists multiple term importance predictors in the area of text retrieval. One option is DeepCT \cite{dai2019context}, which predicts importances for all terms appearing in a given document. However, since DeepCT is not designed to model importances of absent terms, it is not suitable for evaluating absent candidate phrases. In another line of work, EPIC \cite{macavaney2020expansion}, SparTerm \cite{bai2020sparterm} and SPLADE \cite{10.1145/3404835.3463098, https://doi.org/10.48550/arxiv.2109.10086} predict importances for all terms in a vocabulary, making them more suitable for evaluation of both present and absent candidates.

Among these models, SPLADE is the most suitable for predicting keyphrases. Different from EPIC and SparTerm, SPLADE employs explicit sparsity regularization mechanisms, which encourages assigning non-zero importance for only the most relevant terms. In the next sections, we discuss the term importance predictor: SPLADE (\S \ref{section:term_importance_predictor_splade}), training data (\S \ref{section:term_importance_predictor_training_dataset}) and informativeness estimation (\S \ref{section:estimating_informativeness_of_phrases}).

\subsubsection{Term Importance Predictor: SPLADE}
\label{section:term_importance_predictor_splade}

SPLADE predicts term importances for an input document $\boldsymbol{x}$ based on the logits produced by the Masked Language Modeling (MLM) layer. In particular, $w^{\boldsymbol{x}}_{ij}$ denotes the importance, predicted by MLM layer, of the term $i \in \boldsymbol{x}$ and the term $j$ in BERT vocabulary\footnote{In this work, term refers to a word or sub-word in BERT vocabulary}. Then, the importance of $j$ given $\boldsymbol{x}$ is computed by max pooling
\begin{equation}
\label{equation:term_importances_splade}
    w^{\boldsymbol{x}}_j=\max_{i \in \boldsymbol{x}} \log (1 + \text{ReLU}(w^{\boldsymbol{x}}_{ij}))
\end{equation}

\noindent \textbf{Model training}. SPLADE is trained by optimizing a ranking loss and two regularization losses

\vspace{-8pt}

\begin{equation}
    \mathcal{L} = \mathcal{L}_{rank-IBN} + \lambda_{q} \mathcal{L}^{q}_{reg} +\lambda_{d} \mathcal{L}^{d}_{reg} 
\end{equation}
where $\mathcal{L}_{reg}$ is the sparse regularizer introduced in \cite{paria2020minimizing}. Given a training batch, containing the query $q_i$, the positive (referenced) document $d^{+}_i$ and the negative document $d^{-}_i$, the ranking loss $\mathcal{L}_{rank-IBN}$ is a contrastive loss that maximizes the relevance of $d^{+}_i$, while lowering the relevance of $d^{-}_i$. Relevance is measured by dot product between q and d representations from Eq. \ref{equation:term_importances_splade}. For further details, we refer readers to the original papers \cite{10.1145/3404835.3463098, https://doi.org/10.48550/arxiv.2109.10086}.


\subsubsection{Training Dataset}
\label{section:term_importance_predictor_training_dataset}
To train SPLADE, we build a training set $\mathcal{T} = \{(r_i, d^{+}_i, d^{-}_i)\}_{i=1}^{|\mathcal{T}|}$, containing triplets, where $r_i$ is a reference, while $d^{+}_i$ and $d^{-}_i$ denote positive (referenced) and negative documents, respectively. We note that references $r_i$ are used in place of queries $q_i$. In this work, we focus on scientific text, as all three types of references (queries, citation contexts, and titles) are readily accessible in this domain. 



\noindent \textbf{Query}. Our work leverages training data from the Search task within SciRepEval \footnote{\url{https://huggingface.co/datasets/allenai/scirepeval/viewer/search}} \cite{singh-etal-2023-scirepeval}, which contains about 478k queries from real users on Semantic Scholar. Each query accompanies a list of candidates and their relevance score. We build triplets from this dataset by regarding query as reference $r_i$. We concatenate the title and abstract of each candidate as $d^{+}_i$ for those with a relevance score > 0, and as $d^{-}_i$ for those with a relevance score = 0.

\noindent \textbf{Citation context}. We utilize the permissively licensed subset of unarXive\footnote{\url{https://zenodo.org/records/7752615}} \cite{Saier2023unarXive}, which contains over 165k full-text documents. For each document, we extract citing sentences as references $r_i$. We employ only citation contexts that cite one paper, or collectively cite several paper as a single group, to ensure focus on the concepts of the referenced document. The concatenated titles and abstracts of cited articles are chosen to be positive documents $d^{+}_i$. Negative documents $d^{-}_i$ are similarly constructed by concatenating titles and abstracts but are selected from research articles cited in different sections of the same paper, distinct from the section containing the citing sentence.


\noindent \textbf{Title}. We continue to utilize unarXive dataset \cite{Saier2023unarXive}. More specifically, we designate titles as $r_i$ and corresponding abstracts as $d^{+}_i$ for research articles. For negative documents $d^{-}_i$, we select abstracts of other research articles cited within the paper.



\subsubsection{Estimating Informativeness of Phrases}
\label{section:estimating_informativeness_of_phrases}

In this section, we explain how to measure informativeness of phrases based on the term importance predictor described above. A simple approach is to aggregate the importance of the component terms. More formally, the probability that a candidate phrase $\boldsymbol{c}$ is informative given the document $\boldsymbol{x}$ is defined as

\vspace{-8pt}

\begin{equation}
\label{equation:informativeness_distribution_first_version}
    P_{\text{in}}(\boldsymbol{c}|\boldsymbol{x}) \propto f(\boldsymbol{c}, \boldsymbol{x})= \frac{1}{|\boldsymbol{c}| - \gamma} \sum_{c_i \in \boldsymbol{c}} w^{\boldsymbol{x}}_{c_i}
\end{equation}

where $w^{\boldsymbol{x}}_{c_i}$ is the predicted importance of term \( c_i \in \boldsymbol{c} \). Next, $\gamma$ is the length penalty, which is used to control the preference towards longer candidates. A negative value of $\gamma$ leads to larger value of $f(\boldsymbol{c}, \boldsymbol{x})$ for longer candidates, and vice versa.

Although SPLADE can evaluate importance of absent terms, the scores for these terms are often underestimated. On a set of 20k documents sampled from SciRepEval Search, only 25\% of terms with non-zero importances are absent terms. This could lead to inaccurate ranking of absent candidates. To mitigate this problem, our approach is inspired by pseudo-relevance feedback \cite{cao2008selecting}, which is to incorporate additional context from related documents. In particular, the importance of each candidate is determined by its importance in the given document $\boldsymbol{x}$ and its related documents $\boldsymbol{x}' \in \mathcal{N}(\boldsymbol{x})$. Consequently, the informativeness probability is redefined as follows

\vspace{-8pt}

\begin{equation}
\label{equation:informativeness_distribution_second_version}
P_{\text{in}}(\boldsymbol{c}|\boldsymbol{x}) \propto \hat{f}(\boldsymbol{c}, \boldsymbol{x})= \frac{1}{|\boldsymbol{c}| - \gamma} \sum_{c_i \in \boldsymbol{c}} \hat{w}^{\boldsymbol{x}}_{c_i}
\end{equation}

\begin{equation}
\label{equation:informativeness_distribution_second_version_2nd_part}
    \hat{w}^{\boldsymbol{x}}_{c_i}=\alpha \ w^{\boldsymbol{x}}_{c_i} + (1-\alpha) \sum_{\boldsymbol{x}' \in \mathcal{N}(\boldsymbol{x})}\tilde{s}_{\boldsymbol{x}',\boldsymbol{x}} \ w^{\boldsymbol{x}'}_{c_i}
\end{equation}


 Here, $\mathcal{N}(\boldsymbol{x})$ is retrieved using BM25 from a document collection $\mathcal{D}$. The hyperparameter $\alpha$ controls the relative contribution of the given document and its related documents. $\tilde{s}_{\boldsymbol{x}',\boldsymbol{x}}=\frac{s_{\boldsymbol{x}',\boldsymbol{x}}}{\sum_{\boldsymbol{x}'' \in \mathcal{N}(\boldsymbol{x})} s_{\boldsymbol{x}'',\boldsymbol{x}}}$ is the normalized similarity between two documents, where $s_{\boldsymbol{x}',\boldsymbol{x}}$ denotes the BM25 similarity score.  It is worth noting that the term importances of the documents in $\mathcal{D}$ are precomputed and therefore no additional computations are required. 

\subsection{Phraseness Module}

The phraseness module is responsible for generating keyphrase candidates, including present and absent ones. A discussion in \cite{do-etal-2023-unsupervised} mentions that most keyphrases are noun phrases \cite{chuang2012without} and absent keyphrases can be found in other documents \cite{ye-etal-2021-heterogeneous}. Based on this idea, we employ a candidate generation procedure that extract noun phrases from 1) the given document $\boldsymbol{x}$ and 2) its related documents $\mathcal{N}(\boldsymbol{x})$. More formally, given a document, its candidate set $\hat{\boldsymbol{C}}_{\boldsymbol{x}} = \{\boldsymbol{c}_1, \boldsymbol{c}_2,...\}$ containing keyphrase candidates, is obtained as follows

\vspace{-8pt}

\begin{equation}
\label{equation:candidate_set_formation}
\begin{split}
    \hat{\boldsymbol{C}}_{\boldsymbol{x}} &= \boldsymbol{C}_{\boldsymbol{x}} \ \cup \boldsymbol{C}_{\mathcal{N}(\boldsymbol{x})} =\boldsymbol{C}_{\boldsymbol{x}} \ \cup \bigcup_{\boldsymbol{x}' \in \mathcal{N}(\boldsymbol{x})} \boldsymbol{C}_{\boldsymbol{x}'}
\end{split}
\end{equation}

where $\boldsymbol{C}_{\boldsymbol{x}}$ denotes the set of noun phrases extracted from $\boldsymbol{x}$. $\boldsymbol{C}_{\mathcal{N}(\boldsymbol{x})}$ denotes the union of noun phrase sets $\boldsymbol{C}_{\boldsymbol{x}'}$ from related documents $\boldsymbol{x}' \in \mathcal{N}(\boldsymbol{x})$. To assign a phraseness probability of each candidate $\boldsymbol{c} \in \hat{\boldsymbol{C}}_{\boldsymbol{x}}$, we compute the likelihood that it is drawn from either the noun phrases in the given document ($\boldsymbol{C}_{\boldsymbol{x}}$) or those in the related documents ($\boldsymbol{C}_{\boldsymbol{x}'}$)
\vspace{-8pt}

\begin{equation}
\begin{split}
\label{equation:phraseness_distribution}
    &P_{\text{pn}}(\boldsymbol{c} | \boldsymbol{x}) = \beta \ P(\boldsymbol{c} | \boldsymbol{C}_{\boldsymbol{x}}) \\ &+(1 - \beta) \sum_{\boldsymbol{x}' \in \mathcal{N(\boldsymbol{x})}} \tilde{s}_{\boldsymbol{x}',\boldsymbol{x}} \ P(\boldsymbol{c} | \boldsymbol{C}_{\boldsymbol{x}'})
\end{split}
\end{equation}


\begin{equation}
    P(\boldsymbol{c} | \boldsymbol{C})=\begin{cases}
			\frac{1}{|\boldsymbol{C}|}, & \boldsymbol{c} \in \boldsymbol{C}\\
            0, & \text{otherwise}
		 \end{cases}
\end{equation}

The parameter $\beta$ controls the relative contribution from the given document and its related documents.

However, as the size of $\mathcal{N}(\boldsymbol{x})$ grows, the size of 
$\boldsymbol{C}_{\mathcal{N}(\boldsymbol{x})}$ (and therefore $\hat{\boldsymbol{C}}_{\boldsymbol{x}}$) may grow significantly. The large number of candidates slows down keyphrase generation process, regardless of how fast informativeness estimation is. To limit the number of candidates for speeding up the KG process, we employ two strategies for pruning $\boldsymbol{C}_{\mathcal{N}(\boldsymbol{x})}$.

\noindent \textbf{Strategy 1}: \textit{Pruning low informativeness and low reliability candidates from each $\boldsymbol{C}_{\boldsymbol{x}'}$}. The informativeness of a candidate given the input document $\boldsymbol{x}$ depends not only on $\boldsymbol{x}$ but also on how important that candidate is to the related documents. Specifically, we can see from Eq. \ref{equation:informativeness_distribution_second_version} and \ref{equation:informativeness_distribution_second_version_2nd_part} that unimportant candidates given the related documents are likely to have low informativeness and hence unlikely to be chosen as keyphrases. Based on this idea, we prune $\boldsymbol{C}_{\boldsymbol{x}'}$ by keeping only the top 10 $\boldsymbol{c} \in \boldsymbol{C}_{\boldsymbol{x}'}$ with the highest value of $f(\boldsymbol{c}, \boldsymbol{x}')$ (see Eq. \ref{equation:informativeness_distribution_first_version}).

Next, we further prune $\boldsymbol{C}_{\boldsymbol{x}'}$ based on their \textit{reliability}. Inspired by \cite{boudin-aizawa-2024-unsupervised}, we estimate phrase reliability by using the number of documents in which they appear as one of the most informative candidates. Intuitively, a phrase is reliable if it is used frequently to describe key concepts. With this in mind, we employ $\boldsymbol{G}_{\mathcal{D}}$, which is a glossary formed by retaining noun phrases that appear in the top 10 most informative candidates for at least three documents $\boldsymbol{x}' \in \mathcal{D}$. Applying the first strategy, the pruned candidate set from related documents $\boldsymbol{x}'$, denoted as $\tilde{\boldsymbol{C}}_{\boldsymbol{x}'}$, is defined as follows

\vspace{-8pt}

\begin{equation}
    \tilde{\boldsymbol{C}}_{\boldsymbol{x}'}=\text{Top}_{10}(\boldsymbol{C}_{\boldsymbol{x}'}, f) \cap \boldsymbol{G}_{\mathcal{D}} 
\end{equation}

We note that $\tilde{\boldsymbol{C}}_{\boldsymbol{x}'}$ is precomputed for every document in $\mathcal{D}$ and therefore no additional computations are required in the inference phase. The pruned candidate sets $\tilde{\boldsymbol{C}}_{\boldsymbol{x}'}$ are used in place of $\boldsymbol{C}_{\boldsymbol{x}'}$ in Eq. \ref{equation:candidate_set_formation} and \ref{equation:phraseness_distribution}.

\noindent \textbf{Strategy 2}: \textit{Pruning low phraseness candidates from $\boldsymbol{C}_{\mathcal{N}(\boldsymbol{x})}$}. As will be discussed in \S \ref{section:combining_phraseness_and_informativeness}, candidates chosen as keyphrases not only need to exhibit high informativeness, but also phraseness probability. Therefore, candidates with low phraseness are unlikely to be chosen as keyphrases. Based on this idea, we prune $\boldsymbol{C}_{\mathcal{N}(\boldsymbol{x})}$ by retaining only the top 100 with the highest value of $P_{\text{pn}}(\boldsymbol{c} | \boldsymbol{x})$. Applying the second strategy, the final candidate set is redefined as follows

\vspace{-8pt}
\begin{equation}
    \hat{\boldsymbol{C}}_{\boldsymbol{x}} = \boldsymbol{C}_{\boldsymbol{x}} \ \cup \text{Top}_{100}(\boldsymbol{C}_{\mathcal{N}(\boldsymbol{x})}, P_{\text{pn}})
\end{equation}

\subsection{Combining Phraseness and Informativeness}
\label{section:combining_phraseness_and_informativeness}

To generate keyphrases, we combine the two modules. Specifically, given an input text, we first apply the phraseness module to generate keyphrase candidates $\hat{\boldsymbol{C}}_{\boldsymbol{x}}$. Next, we evaluate the informativeness of each candidate. The candidates are ranked based on a composite ranking score, which is computed as the product-of-experts \cite{hinton2002training} of the phraseness and informativeness probabilities

\vspace{-8pt}

\begin{equation}
\label{equation:keyphrase_distribution}
    P_{\text{kp}}(\boldsymbol{c} | \boldsymbol{x}) \propto P_{\text{pn}}(\boldsymbol{c} | \boldsymbol{x}) ^ {\lambda} \times P_{\text{in}}(\boldsymbol{c} | \boldsymbol{x})
\end{equation}

where $\lambda$ is a hyperparameter that controls the importance of phraseness in the ranking score.

\noindent \textbf{Position penalty}. Previous work have shown that position information is useful for predicting present keyphrases (i.e. keyphrase extraction) \cite{florescu-caragea-2017-positionrank, boudin-2018-unsupervised, gallina2020large}. Therefore, we include this feature into measuring informativeness of phrases. In particular, we adopt the position penalty defined in \cite{do-etal-2023-unsupervised}. The final ranking score is defined as follows

\begin{equation}
\label{equation:final_ranking_function}
    \boldsymbol{s}_{\boldsymbol{x}}(\boldsymbol{c}) = \omega_{\boldsymbol{x}}(\boldsymbol{c}) P_{\text{kp}}(\boldsymbol{c} | \boldsymbol{x})
\end{equation}

where $\omega_{\boldsymbol{x}}(\boldsymbol{c}) =1+\frac{1}{\log_2[\mathcal{P}_{\boldsymbol{x}}(\boldsymbol{c}) + 2]}$ is the position penalty. The position $\mathcal{P}_{\boldsymbol{x}}(\boldsymbol{c})$ is the number of words preceding the phrase  $\boldsymbol{c}$ in $\boldsymbol{x}$. This penalty prioritizes phrases appearing earlier in the text. For absent phrases, we define $\mathcal{P}_{\boldsymbol{x}}(\boldsymbol{c}) \rightarrow \infty$ and therefore $\omega_{\boldsymbol{x}}(\boldsymbol{c}) \rightarrow 1$. Finally, top ranked (present or absent) candidates are chosen as (present or absent) keyphrases.

\noindent \textbf{Switching between generation and extraction}. Our proposed framework can flexibly switch between generation and extraction. This is achieved by setting both interpolation hyperparameters, $\alpha$ and $\beta$ (Eq. \ref{equation:informativeness_distribution_second_version_2nd_part} and \ref{equation:phraseness_distribution}, respectively) to 1. Setting these two parameters to 1 disables the use of $\mathcal{N}(\boldsymbol{x})$ and therefore is equivalent to not retrieving any related documents, i.e. $|\mathcal{N}(\boldsymbol{x})|=0$.


\section{Experiments}

\begin{table*}[h]
\centering
\resizebox{0.9\textwidth}{!}{
\begin{tabular}{|lllllllllllll|}
\hline
\multicolumn{13}{|c|}{\textbf{Present keyphrase generation}} \\ \hline
\multicolumn{1}{|l|}{} &
  \multicolumn{2}{c|}{SemEval} &
  \multicolumn{2}{c|}{Inspec} &
  \multicolumn{2}{c|}{NUS} &
  \multicolumn{2}{c|}{Krapivin} &
  \multicolumn{2}{c|}{KP20K} &
  \multicolumn{2}{c|}{Avg} \\
\multicolumn{1}{|l|}{\multirow{-2}{*}{}} &
  \multicolumn{1}{c}{F@5} &
  \multicolumn{1}{c|}{F@10} &
  \multicolumn{1}{c}{F@5} &
  \multicolumn{1}{c|}{F@10} &
  \multicolumn{1}{c}{F@5} &
  \multicolumn{1}{c|}{F@10} &
  \multicolumn{1}{c}{F@5} &
  \multicolumn{1}{c|}{F@10} &
  \multicolumn{1}{c}{F@5} &
  \multicolumn{1}{c|}{F@10} &
  \multicolumn{1}{c}{F@5} &
  \multicolumn{1}{c|}{F@10} \\ \hline
\multicolumn{1}{|l|}{TextRank} &
  16 &
  \multicolumn{1}{l|}{20.3} &
  29.3 &
  \multicolumn{1}{l|}{36.2} &
  11.6 &
  \multicolumn{1}{l|}{16.6} &
  10.1 &
  \multicolumn{1}{l|}{13.6} &
  9.1 &
  \multicolumn{1}{l|}{11.6} &
  15.2 &
  19.7 \\
\multicolumn{1}{|l|}{MultiPartiteRank} &
  22.3 &
  \multicolumn{1}{l|}{22.5} &
  26.3 &
  \multicolumn{1}{l|}{30.3} &
  23.7 &
  \multicolumn{1}{l|}{22.2} &
  17.9 &
  \multicolumn{1}{l|}{15.9} &
  18.4 &
  \multicolumn{1}{l|}{15.9} &
  21.7 &
  21.4 \\
\multicolumn{1}{|l|}{EmbedRank} &
  23.5 &
  \multicolumn{1}{l|}{25.2} &
  27.9 &
  \multicolumn{1}{l|}{33.4} &
  23.8 &
  \multicolumn{1}{l|}{22.3} &
  18.6 &
  \multicolumn{1}{l|}{17.7} &
  19.5 &
  \multicolumn{1}{l|}{16.8} &
  22.7 &
  23.1 \\
\multicolumn{1}{|l|}{EmbedRank (SBERT)} &
  25.4 &
  \multicolumn{1}{l|}{27.1} &
  \textbf{35.1} &
  \multicolumn{1}{l|}{\textbf{39.8}} &
  22.5 &
  \multicolumn{1}{l|}{24.1} &
  20.7 &
  \multicolumn{1}{l|}{19.3} &
  18.3 &
  \multicolumn{1}{l|}{17.1} &
  24.4 &
  25.5 \\
\multicolumn{1}{|l|}{PromptRank} &
  16.1 &
  \multicolumn{1}{l|}{19.9} &
  33.4 &
  \multicolumn{1}{l|}{{\ul 37.5}} &
  18.5 &
  \multicolumn{1}{l|}{19.8} &
  15.9 &
  \multicolumn{1}{l|}{15.5} &
  16.3 &
  \multicolumn{1}{l|}{15.6} &
  20 &
  21.7 \\ \hline
\multicolumn{1}{|l|}{AutoKeyGen} &
  22.1 &
  \multicolumn{1}{l|}{24.4} &
  23.1 &
  \multicolumn{1}{l|}{23.7} &
  26.1 &
  \multicolumn{1}{l|}{\textbf{27.1}} &
  20.6 &
  \multicolumn{1}{l|}{18.6} &
  20.4 &
  \multicolumn{1}{l|}{19} &
  22.5 &
  22.6 \\
\multicolumn{1}{|l|}{UOKG} &
  21.5 &
  \multicolumn{1}{l|}{22.1} &
  23.9 &
  \multicolumn{1}{l|}{22.9} &
  {\ul 27.8} &
  \multicolumn{1}{l|}{26.2} &
  \textbf{21.5} &
  \multicolumn{1}{l|}{17.9} &
  21 &
  \multicolumn{1}{l|}{17.6} &
  23.1 &
  21.3 \\
\multicolumn{1}{|l|}{TPG} &
  24.7 &
  \multicolumn{1}{l|}{22.2} &
  {\ul 34} &
  \multicolumn{1}{l|}{33.3} &
  25 &
  \multicolumn{1}{l|}{21.3} &
  20.3 &
  \multicolumn{1}{l|}{16.3} &
  18.7 &
  \multicolumn{1}{l|}{14.2} &
  24.5 &
  21.5 \\ \hline
\multicolumn{1}{|l|}{\textbf{ERU-KG-small}} &
  {\ul 27.4}$^*$ &
  \multicolumn{1}{l|}{{\ul 30.1}$^*$} &
  28.4 &
  \multicolumn{1}{l|}{35.7} &
  \textbf{28.1}$^*$ &
  \multicolumn{1}{l|}{26.9} &
  20.7 &
  \multicolumn{1}{l|}{\textbf{19.6}} &
  {\ul 21.6}$^*$ &
  \multicolumn{1}{l|}{{\ul 19.2}} &
  {\ul 25.2} &
  {\ul 26.3}$^*$ \\
\multicolumn{1}{|l|}{\textbf{ERU-KG-base}} &
  \textbf{27.6}$^*$ &
  \multicolumn{1}{l|}{\textbf{30.6}$^*$} &
  29 &
  \multicolumn{1}{l|}{36} &
  {\ul 27.8} &
  \multicolumn{1}{l|}{{\ul 27}} &
  {\ul 21.3} &
  \multicolumn{1}{l|}{{\ul 19.5}$^*$} &
  \textbf{22}$^*$ &
  \multicolumn{1}{l|}{\textbf{19.4}$^*$} &
  \textbf{25.5} &
  \textbf{26.5}$^*$ \\ \hline
\multicolumn{1}{|l|}{Supervised - CopyRNN} &
  29.6 &
  \multicolumn{1}{l|}{29.7} &
  22.6 &
  \multicolumn{1}{l|}{23.7} &
  37.2 &
  \multicolumn{1}{l|}{34.3} &
  30.1 &
  \multicolumn{1}{l|}{24.5} &
  30.6 &
  \multicolumn{1}{l|}{25.7} &
  30 &
  27.6 \\ \hline
\multicolumn{13}{|c|}{\textbf{Absent keyphrase generation}} \\ \hline
\multicolumn{1}{|l|}{} &
  \multicolumn{2}{c|}{SemEval} &
  \multicolumn{2}{c|}{Inspec} &
  \multicolumn{2}{c|}{NUS} &
  \multicolumn{2}{c|}{Krapivin} &
  \multicolumn{2}{c|}{KP20K} &
  \multicolumn{2}{c|}{Avg} \\
\multicolumn{1}{|l|}{\multirow{-2}{*}{}} &
  \multicolumn{1}{c}{R@5} &
  \multicolumn{1}{c|}{R@10} &
  \multicolumn{1}{c}{R@5} &
  \multicolumn{1}{c|}{R@10} &
  \multicolumn{1}{c}{R@5} &
  \multicolumn{1}{c|}{R@10} &
  \multicolumn{1}{c}{R@5} &
  \multicolumn{1}{c|}{R@10} &
  \multicolumn{1}{c}{R@5} &
  \multicolumn{1}{c|}{R@10} &
  \multicolumn{1}{c}{R@5} &
  \multicolumn{1}{c|}{R@10} \\ \hline
\multicolumn{1}{|l|}{AutoKeyGen} &
  0.7 &
  \multicolumn{1}{l|}{1.1} &
  1.8 &
  \multicolumn{1}{l|}{2.6} &
  2.3 &
  \multicolumn{1}{l|}{3.2} &
  2.5 &
  \multicolumn{1}{l|}{3.7} &
  2.2 &
  \multicolumn{1}{l|}{3.6} &
  1.9 &
  2.8 \\
\multicolumn{1}{|l|}{UOKG} &
  1.4 &
  \multicolumn{1}{l|}{2.3} &
  1.9 &
  \multicolumn{1}{l|}{2.9} &
  2.5 &
  \multicolumn{1}{l|}{3.6} &
  4.6 &
  \multicolumn{1}{l|}{\textbf{6.9}} &
  2.6 &
  \multicolumn{1}{l|}{4.5} &
  2.6 &
  4 \\
\multicolumn{1}{|l|}{TPG} &
  0.4 &
  \multicolumn{1}{l|}{0.8} &
  1.5 &
  \multicolumn{1}{l|}{2.4} &
  1.7 &
  \multicolumn{1}{l|}{2.4} &
  1 &
  \multicolumn{1}{l|}{1.2} &
  1.2 &
  \multicolumn{1}{l|}{1.9} &
  1.2 &
  1.7 \\ \hline
\multicolumn{1}{|l|}{\textbf{ERU-KG-small}} &
  {\ul 2.1}$^*$ &
  \multicolumn{1}{l|}{\textbf{3.1}} &
  \textbf{5.4}$^*$ &
  \multicolumn{1}{l|}{\textbf{6.5}$^*$} &
  \textbf{3.7}$^*$ &
  \multicolumn{1}{l|}{\textbf{5.9}$^*$} &
  \textbf{5} &
  \multicolumn{1}{l|}{{\ul 6.2}} &
  \textbf{6}$^*$ &
  \multicolumn{1}{l|}{{\ul 8}$^*$} &
  \textbf{4.4}$^*$ &
  \textbf{5.9}$^*$ \\
\multicolumn{1}{|l|}{\textbf{ERU-KG-base}} &
  \textbf{2.3}$^*$ &
  \multicolumn{1}{l|}{{\ul 3}$^*$} &
  {\ul 5.3}$^*$ &
  \multicolumn{1}{l|}{\textbf{6.5}$^*$} &
  {\ul 3.4}$^*$ &
  \multicolumn{1}{l|}{{\ul 5.5}$^*$} &
  {\ul 4.9} &
  \multicolumn{1}{l|}{{\ul 6.2}} &
  \textbf{6}$^*$ &
  \multicolumn{1}{l|}{\textbf{8.1}$^*$} &
  \textbf{4.4}$^*$ &
  {\ul 5.8}$^*$ \\ \hline
\multicolumn{1}{|l|}{Supervised - CopyRNN} &
  2.3 &
  \multicolumn{1}{l|}{2.8} &
  3.5 &
  \multicolumn{1}{l|}{4.9} &
  5.9 &
  \multicolumn{1}{l|}{7.8} &
  7.9 &
  \multicolumn{1}{l|}{10.8} &
  7.1 &
  \multicolumn{1}{l|}{9.3} &
  5.3 &
  7.1 \\ \hline
\end{tabular}
}
\caption{Keyphrase generation performances on five benchmark datasets. The best results are bolded, while the second-best are underlined. Experiments for AutoKeyGen, UOKG, TPG, CopyRNN, and our method are conducted three times, with the mean reported. Both F1 and Recall are presented as percentages. $^*$ indicates significance
over AutoKeyGen, UOKG and TPG with $p<0.05$.}
\label{table:keyphrase_generation_performance}
\end{table*}

In this work, we assess the effectiveness of ERU-KG using two evaluation methods: \textbf{Ground truth-based} and \textbf{Retrieval-based} evaluation. The former measures the alignment between predicted keyphrases and human-annotated keyphrases, while the latter assesses the usefulness of predicted keyphrases when applied to text retrieval tasks. More specifically, retrieval-based evaluation aim to determine if keyphrases effectively serve as query and document expansion to enhance text retrieval performance. The datasets, baselines \& evaluation metrics, and experiment results are respectively presented in \S \ref{section:experiments_datasets}, \S \ref{section:experiments_baselines_eval_metrics} and \S \ref{section:experiments_results}.



One of the core contributions of this work is that keyphrase generation can be made more \textit{time-efficient} by leveraging term-based representations of documents. To validate this, we conduct \textbf{Inference speed} evaluation (\S \ref{section:inference_speed_evaluation}). 


\subsection{Datasets}

\label{section:experiments_datasets}

We present the statistics for the evaluation datasets in Table \ref{table:evaluation_datasets_statistics}.

\textbf{Ground truth-based evaluation}. We utilize 5 datasets, namely \textit{SemEval} \cite{kim-etal-2010-semeval}, \textit{Inspec} \cite{hulth-2003-improved}, \textit{NUS} \cite{nguyen2007keyphrase}, \textit{Krapivin} \cite{krapivin2009large} and \textit{KP20K} \cite{meng-etal-2017-deep} for the ground truth-based evaluation of our model. We follow previous work and form the testing document by concatenating the title and abstract of each testing example. 

\noindent \textbf{Retrieval-based evaluation}. We utilize 6 scientific retrieval datasets. Four of these datasets - \textit{TREC-COVID} \cite{voorhees2021trec}, \textit{SCIDOCS} \cite{cohan-etal-2020-specter}, \textit{SciFact} \cite{wadden-etal-2020-fact} and \textit{NFCorpus} \cite{boteva2016full} - are sourced from the BEIR benchmark \cite{thakur2021beir}. The other two datasets are \textit{DORIS-MAE} \cite{wang2024scientific} and \textit{ACM-CR} \cite{boudin2021acm}.

\begin{table*}[h]
\centering
\resizebox{0.85\textwidth}{!}{

\begin{tabular}{|l|l|cccccc|c|}
\hline
\multicolumn{1}{|c|}{Type} & \multicolumn{1}{c|}{Model} & SCIDOCS  & SciFact      & TREC-COVID    & NFCorpus      & DORIS-MAE  & ACM-CR        & Avg           \\ \hline
-                         & BM25                    & 56.4          & 97.7       & 39.6       & 37       & 70.1          & 71.5       & 62.1     
\\ \hline
\multirow{7}{*}{Query}     & + RM3                      & {\ul 59} & 98           & {\ul 44.5}    & \textbf{56.5} & 59.6       & \textbf{74.4} & 65.3          \\
                          & + AutoKeyGen            & 52.3          & 97         & 33.4       & 48.7     & 70.4          & 69.2       & 61.8       \\
                          & + UOKG                  & 54.2          & 98         & 35.4       & 48.6     & 69            & 70.2       & 62.6       \\
                          & + TPG                   & 54.1          & 98.3       & 34.5       & 48.1     & \textbf{73.9} & 71         & 63.3       \\
                          & + CopyRNN               & 53.6          & 97.7       & 35.8       & 48       & 72.8          & 73.8       & 63.6       \\
                          & \textbf{+ ERU-KG-small} & 58.5          & {\ul 99.3} & 43.7       & 56.3     & \textbf{73.9} & 72.1       & {\ul 67.3} \\
                          & \textbf{+ ERU-KG-base}  & 58.7          & 99         & 43.2       & 54.8     & 73.4          & 72.6       & 67         \\ \hline
\multirow{6}{*}{Document} & + docT5query            & 57            & 98         & {\ul 43.2} & 37       & -             & -          & -          \\
                          & + AutoKeyGen            & 57            & 97.3       & 40.5       & 37.3     & 69.8          & 71.3       & 62.2       \\
                          & + UOKG                  & 57.7          & 97.7       & 40.9       & 37.5     & {\ul 70.1}    & 72.4       & 62.7       \\
                          & + CopyRNN               & 57            & 97.3       & 40.8       & 37.2     & 69.7          & 71.6       & 62.3       \\
                          & \textbf{+ ERU-KG-small} & 59.9          & {\ul 98.3} & 38.5       & {\ul 39} & 68.9          & {\ul 73}   & {\ul 62.9} \\
                          & \textbf{+ ERU-KG-base}  & {\ul 60}      & {\ul 98.3} & 39.6       & 38.7     & 68            & 72.7       & {\ul 62.9} \\ \hline
\multirow{6}{*}{Both}      & + docT5query + RM3         & 59.7     & 98.3         & \textbf{47.7} & \textbf{56.5} & -          & -             & -             \\
                          & + AutoKeyGen            & 52.8          & 97         & 33.5       & 48.3     & 69.3          & 68         & 61.5       \\
                          & + UOKG                  & 54.8          & 98.3       & 36.1       & 49.2     & 69.1          & 69.4       & 62.8       \\
                          & + CopyRNN               & 54.7          & 97.5       & 32.4       & 48.1     & 72            & {\ul 73.8} & 63.1       \\
                           & \textbf{+ ERU-KG-small}    & 62.4     & \textbf{100} & 46.2          & 56.2          & {\ul 73.6} & 72.8          & \textbf{68.5} \\
                          & \textbf{+ ERU-KG-base}  & \textbf{62.9} & 99.7       & 46.7       & 55.6     & 71.7          & 73.5       & 68.4       \\ \hline
\end{tabular}

}
\caption{Retrieval-based evaluation (\textbf{R@1000}) on four benchmark datasets, reported as percentages. For each dataset, we \textbf{bold} the best overall results and \underline{underline} the best results in each type (query expansion, document expansion and both).}
\label{table:retrieval_based_evaluation}
\end{table*}

\subsection{Baselines \& Evaluation Metrics}

\label{section:experiments_baselines_eval_metrics}
\subsubsection{Baselines} 

\textbf{Ground truth-based evaluation}. We evaluate our proposed model by comparing against four unsupervised keyphrase extraction algorithms: TextRank \cite{mihalcea-tarau-2004-textrank}, MultiPartiteRank \cite{boudin-2018-unsupervised}, EmbedRank \cite{bennani-smires-etal-2018-simple}, and PromptRank \cite{kong-etal-2023-promptrank}.

Additionally, we compare our model with three unsupervised keyphrase generation methods: AutoKeyGen \cite{shen2022unsupervised}, UOKG \cite{do-etal-2023-unsupervised} and TPG (zero-shot setting) \cite{kang-shin-2024-improving}. Finally, we include CopyRNN \cite{meng-etal-2017-deep} as a supervised baseline. 


\noindent \textbf{Retrieval-based evaluation}. We compare ERU-KG with keyphrase generation methods mentioned above. For all keyphrase generation models, we generate keyphrases for each document (and/or query). We employ the top 10 present keyphrases and top 10 absent keyphrases (20 total) as query and document expansions. In the case of TPG, we evaluate its performance solely on query expansion, due to its slow inference speed. In addition, we compare our model with well-established methods, specifically DocT5Query \cite{nogueira2019document, nogueira2019doc2query} for document expansion and RM3 \cite{abdul2004umass} for query expansion. All expansion techniques are followed by BM25 retrieval.

\subsubsection{Evaluation Metrics}
\textbf{Ground truth-based evaluation}. In line with previous work, we utilize the macro-average F1-score and Recall for evaluation of present and absent keyphrases. For both, we conduct evaluations at top 5 and 10 predictions. Before evaluation, both the predicted and ground truth keyphrases are processed using Porter Stemmer \cite{porter1980algorithm}, after which duplicates are removed. Our implementation of F1-score is similar to that of \cite{chan-etal-2019-neural}. Specifically, for F1@$k$ we add wrong keyphrases until the number of predictions reaches $k$ if a model predicts fewer than $k$ keyphrases. The purpose of this processing step is to eliminate the favor towards models that produce fewer keyphrases.

\noindent \textbf{Retrieval-based evaluation}. We utilize recall at top 1000 (\textbf{R@1000}) as the primary evaluation metric, with the aim to assess the effectiveness of generated keyphrases in enhancing the recall of First-stage Retrieval.


\subsection{Results}

\label{section:experiments_results}
\subsubsection{Ground truth-based Evaluation}

Table \ref{table:keyphrase_generation_performance} presents the performance of our proposed method and the baselines on the five benchmark datasets. In addition, we report the average performances.

\noindent \textbf{Present keyphrase generation}. For generating present keyphrases, our proposed method achieves the best or second-best performance across all datasets except Inspec. While our model does not outperform the baselines on every dataset, it achieves the highest average results overall. Notably, compared to CopyRNN, a supervised baseline, our model demonstrates competitive results. Specifically, CopyRNN outperforms ERU-KG by only 1.1 percentage point in the overall F1@10 score. This illustrates the effectiveness of our approach, particularly since it is independent of human-labeled keyphrases.

\noindent \textbf{Absent keyphrase generation}. For generating absent keyphrases, our model achieves the best performance across all benchmark datasets, leading to the highest average results overall. Furthermore, our approach continues to demonstrate competitive performance in comparison to the supervised baseline.

\subsubsection{Retrieval-based Evaluation}

Table \ref{table:retrieval_based_evaluation} displays the performance of our model and the baselines on six text retrieval evaluation datasets. In addition, we report average performance across datasets. For KG models, we investigate their effectiveness in three settings: 1) when employed as query expansion (\textit{Query}); 2) when employed as document expansion (\textit{Doc}) and 3) when employed as both query and document expansion (\textit{Both}). 

\noindent \textbf{Comparison with KG methods}. In the \textit{Query} and \textit{Both} setting, ERU-KG consistently achieve the best performance among existing KG models across datasets, with one exception being the \textit{ACM-CR} dataset, where ERU-KG is second best after CopyRNN. In the \textit{Doc} setting, the performance gain is less consistent. In particular, although our proposed method achieves performance that matches or exceeds the baselines on the majority of datasets, it is outperformed by all baselines on \textit{TREC-COVID} and \textit{DORIS-MAE}.

In addition, it is worth noting that when employed as query and document expansion in conjunction (i.e. \textit{Both} setting), ERU-KG on average results in superior performance comparing to \textit{Query} and \textit{Doc} setting, where query and document expansion are employed individually. This effect is not evident in other KG models.

\noindent \textbf{Comparison with existing expansion methods}. ERU-KG achieves performance on par with RM3 in the \textit{Query} setting, DocT5Query in the \textit{Doc} setting, and DocT5Query + RM3 in the \textit{Both} setting. While it does not demonstrate a clear performance advantage over existing expansion methods, it offers a distinct benefit in terms of visualizability. Specifically, the keyphrases generated by ERU-KG are more structured and concise, making them easier to visualize compared to the term-based expansions of RM3 and the synthetic queries produced by DocT5Query.




\subsection{Inference Speed Evaluation}
\label{section:inference_speed_evaluation}

We evaluate the inference speed of our method to measure its time efficiency. Throughput (\textbf{TP}), defined as the number of documents processed per second, serves as the primary metric for this assessment.
ERU-KG is tested in two scenarios: \textit{keyphrase extraction} and \textit{keyphrase generation}. In the \textit{keyphrase extraction} scenario, we compare ERU-KG ($\alpha$ and $\beta$ set to 1, as described in \S \ref{section:combining_phraseness_and_informativeness}) against EmbedRank and PromptRank, using SBERT in place of Sent2vec for EmbedRank to ensure a fair comparison. For the \textit{keyphrase generation} scenario, we benchmark ERU-KG against the previously mentioned KG baselines, along with an additional baseline, PromptKP \cite{wu2022fast} — a non-autoregressive supervised keyphrase generation model. Furthermore, we evaluate two configurations of ERU-KG by varying the size of $\mathcal{N}(\boldsymbol{x})$, setting it to 100 (default), 50 and 10. For fair comparison, we run all experiments with batch size of 1, on the same hardware (see \S \ref{section:computing_infrastructure}), using a dataset composed of SemEval, Inspec, NUS and Krapivin.

We present the results in Table \ref{table:inference_speed}. ERU-KG achieves the best throughput in both scenarios. Results in the keyphrase generation scenario requires further explanations. In the default setting, i.e. $|\mathcal{N}(\boldsymbol{x})|=100$, our proposed method fails to achieve a clear advantage over all baselines. However, when setting $|\mathcal{N}(\boldsymbol{x})|$ to smaller sizes, e.g. 50 or 10, ERU-KG becomes significantly faster. This shows that the retrieval of related documents is the bottleneck and create a trade-off between effectiveness and efficiency, as will be illustrated in \S \ref{section:ablation_studies:size_of_related_documents_set}, retrieving fewer related documents cause the performance to drop.


\begin{table}[h]
\centering
\resizebox{\columnwidth}{!}{
\begin{tabular}{|c|l|l|cc|}
\hline
Scenario                                                                             & \multicolumn{1}{c|}{Model name} & Note & Model size & TP (doc/s) \\ \hline
\multirow{3}{*}{\begin{tabular}[c]{@{}c@{}}Keyphrase\\ extraction\end{tabular}}  & EmbedRank (SBERT)           & -    & 33M        & {\ul 43.5}        \\
 & PromptRank  & - & 60M  & 1.4           \\
 & \textbf{ERU-KG-base} & $\alpha=1,\beta=1$  & 66M  & \textbf{72.9}$^*$ \\ \hline
\multirow{7}{*}{\begin{tabular}[c]{@{}c@{}}Keyphrase \\ generation\end{tabular}} & AutoKeyGen                      & -    & 37M        & 9.7        \\
 & UOKG        & - & 37M  & 4.8           \\
 & TPG         & - & 139M & 0.8           \\
 & CopyRNN     & - & 37M  & 11          \\
 & PromptKP    & - & 110M & 10.4          \\
 & \textbf{ERU-KG-base} & $|\mathcal{N}(\boldsymbol{x})| = 100$ & 66M  & 10.9          \\
 & \textbf{ERU-KG-base} & $|\mathcal{N}(\boldsymbol{x})|=50$  & 66M  & {\ul 12.1}$^*$ \\
 & \textbf{ERU-KG-base} & $|\mathcal{N}(\boldsymbol{x})|=10$  & 66M  & \textbf{15.5}$^*$ \\ \hline
\end{tabular}
}
\caption{Throughput (\textbf{TP}) of ERU-KG and baselines. We \textbf{bold} and \underline{underline} the highest and second-highest throughput in each scenario. $^*$ denotes significance over the second-best baselines with $p<0.05$, respectively. Statistical significance tests are conducted separately for each scenario.}
\label{table:inference_speed}
\end{table}

\subsection{Ablation Studies}

We conduct two ablation studies to understand 1) how different of references (queries, citation contexts and titles) contribute to ERU-KG performance and 2) how retrieving fewer related documents affect our proposed model's performance. In this section, we conduct the experiments on ERU-KG-base, i.e. the version of ERU-KG with informativeness module initialized from DistilBERT-base. We also provide a sensitivity analysis of $\alpha$ and $\beta$ in \S \ref{section:sensitivity_analysis_interpolation}.

\subsubsection{Contribution of Each Type of References}
\begin{figure}[]
    \centering
    \includegraphics[width=\columnwidth]{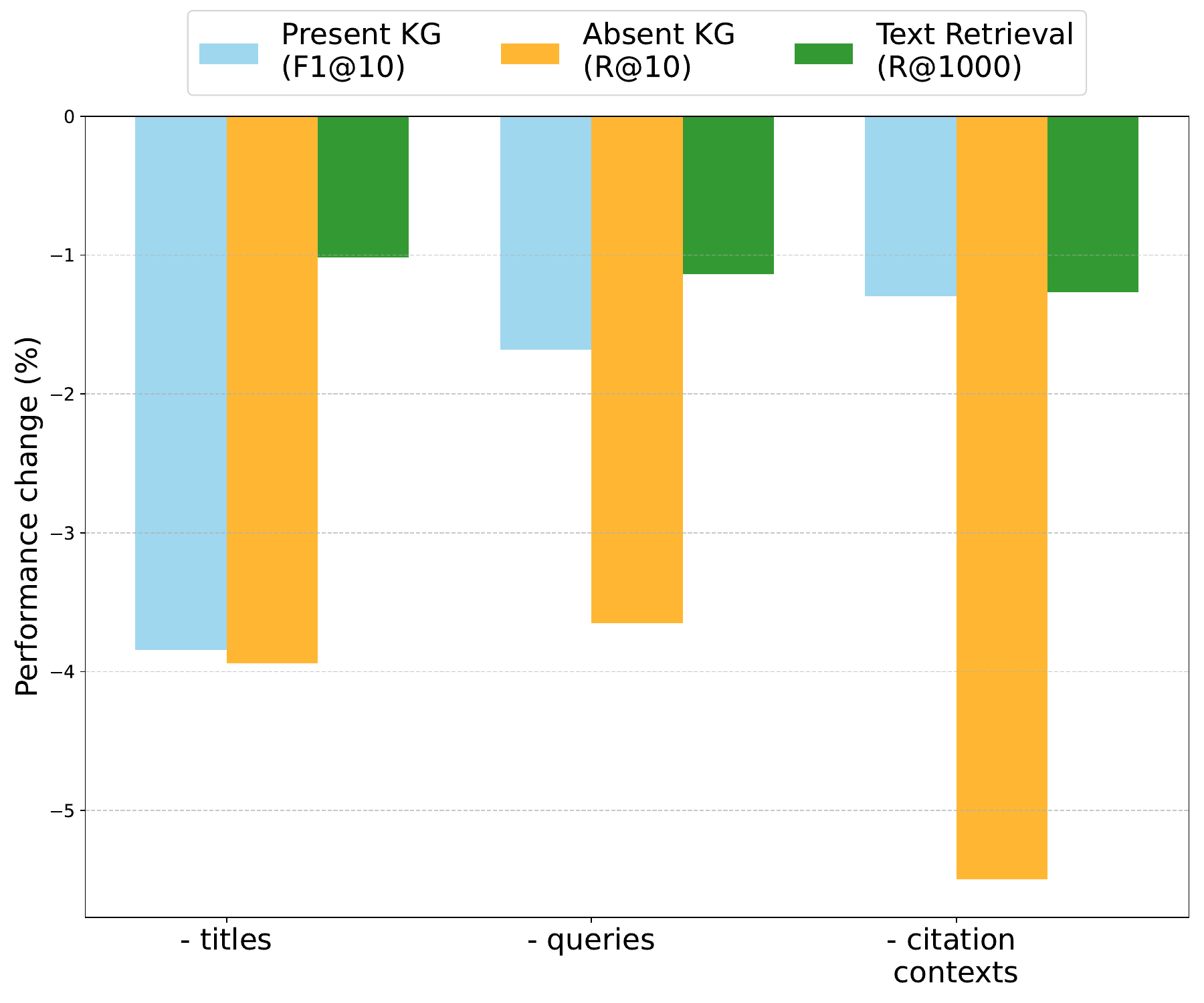}
    \caption{The performance change (in percentage) when excluding one reference type. \textbf{-[type]} indicates the omision of \textbf{[type]}}
    \label{figure:ablation_studies_references_types}
\end{figure}

We study the contribution of each type of references by excluding one type at a time to train variations of ERU-KG. We evaluate the performance change in \textit{keyphrase generation} tasks (F1@10 and R@10 for present and absent keyphrases respectively), \textit{text retrieval} tasks (Recall@1k). We evaluate \textit{text retrieval} in the \textit{Both} setting, where generated keyphrases are used as both query and document expansion. We average the evaluate results across all datasets for each task to measure performance changes. We present the results in Figure \ref{figure:ablation_studies_references_types}. 

For present keyphrase generation (keyphrase extraction), removing title from the training dataset effect performance the most. This suggest that title is a great source of information for enhancing keyphrase extraction, aligning with previous work \cite{chen2019guided, song-etal-2023-improving}. Regarding absent keyphrase generation, performance decreases when any reference type is removed, suggesting that this task is benefitted by understanding how the given document would be mentioned in different contexts. The same comment can be made for text retrieval, where removing any reference type hurt performance.

\subsubsection{Effect of Retrieving Fewer Related Documents}
\label{section:ablation_studies:size_of_related_documents_set}

We study ERU-KG's performance change as it retrieve fewer related documents $\mathcal{N}(\boldsymbol{x})$. Table \ref{table:ablation_studies_num_neighbor} presents the results.

It can be seen that retrieving fewer related documents only affect absent keyphrase generation and text retrieval. Next, we can see that performance gradually decrease as the fewer related documents are retrieved. Notably, when $|\mathcal{N}(\boldsymbol{x})|=10$ the performance drop exceeds 5\% for both absent keyphrase generation and text retrieval. Combining the results with Table \ref{table:inference_speed}, $|\mathcal{N}(\boldsymbol{x})|=50$ appears to strike a good balance between efficiency and effectiveness.

\begin{table}[]
\centering

\resizebox{0.95\columnwidth}{!}{

\begin{tabular}{|c|lll|}
\hline
$|\mathcal{N}(\boldsymbol{x})|$ &
  \multicolumn{1}{c}{\begin{tabular}[c]{@{}c@{}}KG-present\\ (F1@10)\end{tabular}} &
  \multicolumn{1}{c}{\begin{tabular}[c]{@{}c@{}}KG-absent\\ (R@10)\end{tabular}} &
  \multicolumn{1}{c|}{\begin{tabular}[c]{@{}c@{}}TR\\ (R@1000)\end{tabular}} \\ \hline
100 & 26.5 & 5.8 & 68.4 \\
50  & 26.5 & 5.5$\downarrow$ & 67 \\
10  & 26.4 & 4.5$\downarrow$ & 63.1$\downarrow$   \\ \hline
\end{tabular}

}

\caption{The performance change when adjusting the size of related documents set $\mathcal{N}(\boldsymbol{x})$, $\downarrow$ denotes performance drop larger than 5\% in comparison to default setting ($|\mathcal{N}(\boldsymbol{x})| = 100$).}
\label{table:ablation_studies_num_neighbor}
\end{table}

\section{Conclusion}
In this paper, we propose ERU-KG, an unsupervised keyphrase generation model that 1) captures how the community perceives key concepts and 2) estimates informativeness of phrases efficiently. Experiments on keyphrase generation benchmarks demonstrate the effectiveness of ERU-KG. We further validate its performance through evaluations from text retrieval perspective. Notably, the inference speed assessment highlights the model's time efficiency, significantly enhancing its potential for real-world applications.


\section*{Limitations}
In this section, we discuss the limitations of our work. Firstly, we conducted experiments only in the scientific domain, and therefore it is unclear how ERU-KG would perform in other domains. Secondly, we limited our analysis to only three types of references, which may not encompass all possible types (e.g. Tweets referencing a research article). Including additional type of references could improve the performance of our proposed model. Lastly, the design of our phraseness module does not allow customization for absent keyphrase generation. Specifically, since our phraseness module source (absent) keyphrase candidates from other documents, it lacks the flexibility to adapt to the specific context of the given document.





\section*{Acknowledgments}
This material is based upon work supported by the National Science Foundation IIS 16-19302 and IIS 16-33755, Zhejiang University ZJU Research 083650, IBM-Illinois Center for Cognitive Computing Systems Research (C3SR) and IBM-Illinois Discovery Accelerator Institute (IIDAI), grants from eBay and Microsoft Azure, UIUC OVCR CCIL Planning Grant 434S34, UIUC CSBS Small Grant 434C8U, and UIUC New Frontiers Initiative. Any opinions, findings, conclusions, or recommendations expressed in this publication are those of the author(s) and do not necessarily reflect the views of the funding agencies.




\appendix
\begin{table}[h]
\centering
\resizebox{0.85\columnwidth}{!}{
\begin{tabular}{|lccc|}
\hline
\multicolumn{4}{|c|}{\textbf{Groundtruth-based evaluation}}        \\ \hline
\multicolumn{1}{|c|}{Dataset name} & \#doc   & \#kps/doc & \%absent \\ \hline
\multicolumn{1}{|l|}{SemEval}      & 100     & 15.2      & 59.7     \\
\multicolumn{1}{|l|}{Inspec}       & 500     & 9.8       & 22       \\
\multicolumn{1}{|l|}{NUS}          & 211     & 11.6      & 49.3     \\
\multicolumn{1}{|l|}{Krapivin}     & 460     & 5.7       & 51.2     \\
\multicolumn{1}{|l|}{KP20K}        & 19,987  & 5.3       & 44.7     \\ \hline
\multicolumn{4}{|c|}{\textbf{Retrieval-based evaluation}}           \\ \hline
\multicolumn{1}{|c|}{Dataset name} & \#Query & \#Corpus  & Avg D /Q \\ \hline
\multicolumn{1}{|l|}{SCIDOCS}      & 1,000   & 25,657    & 4.9      \\
\multicolumn{1}{|l|}{SciFact}      & 300     & 5,183     & 1.1      \\
\multicolumn{1}{|l|}{TREC-COVID}   & 50      & 171,332   & 493.5    \\
\multicolumn{1}{|l|}{NFCorpus}     & 323     & 3,633     & 38.2     \\
\multicolumn{1}{|l|}{DORIS-MAE}    & 100     & 363,133   & 109.3    \\
\multicolumn{1}{|l|}{ACM-CR}       & 552     & 114,882   & 1.8      \\ \hline
\end{tabular}
}
\caption{Statistics of test splits of evaluation datasets.}
\label{table:evaluation_datasets_statistics}
\end{table}

\section{Related Work}
\label{section:related_work}
\noindent \textbf{Unsupervised keyphrase extraction (UKE)}. UKE focuses on identifying keyphrases within the given text. Previous work typically employ a two-stage procedure: 1) \textit{candidate generation} via ngram or noun phrase extraction; 2) \textit{candidate ranking}, where candidates are ranked based on their informativeness and the top-ranked are selected as keyphrases.

Existing methods in UKE can be classified into four categories, namely \textit{statistics-based}, \textit{graph-based}, 
\textit{embedding-based} and \textit{language model-based}. These categories are distinguished by the importance scoring functions that are used to estimate informativeness, i.e. how candidates are ranked. \textit{Statistics-based} methods \cite{sparck1972statistical, campos2018yake} utilizes features like word frequency, word position, context diversity, etc. \textit{Graph-based} method \cite{mihalcea-tarau-2004-textrank, wan2008single, bougouin2013topicrank, gollapalli2014extracting, florescu-caragea-2017-positionrank, boudin-2018-unsupervised} rank candidates based on different graph-theoretic measures. Embedding-based methods \cite{bennani-smires-etal-2018-simple, sun2020sifrank, zhang-etal-2022-mderank} select candidates that are closest to the given document in the embedding space. \textit{Language model-based} methods utilize Pretrained Language Models (PLMs) to evaluate the informativeness of phrases. \cite{ding-luo-2021-attentionrank} evaluate local and global importance of a candidate by leveraging self and cross attention, \cite{kong-etal-2023-promptrank} estimate informativeness by computing the likelihood of generating the candidate given the input text and a pre-specified prompt.

\noindent \textbf{Unsupervised keyphrase generation (UKG)}. Different from UKE, UKG focuses on generating both present and absent keyphrases. Similar to UKE methods, UKG models typically rely on importance scoring functions, but they are utilized in two distinct ways: 1) to extract silver-labeled data for training seq2seq models or 2) to guide the generation of noun phrases towards those that represent the core concepts. 

The first approach is exemplified by AutoKeyGen \cite{shen2022unsupervised} and Title Phrase Generation (TPG) \cite{kang-shin-2024-improving}. AutoKeyGen trains a seq2seq model on silver-labeled data, where present keyphrases are sourced directly from the text, and absent keyphrases are synthesized by combining present terms. To select present and absent keyphrases, AutoKeyGen employ an importance score that combine semantic and lexical similarity between keyphrase candidates and the document. TPG proposes extracting phrases from titles as silver-labeled keyphrases to train a seq2seq model. 

The second approach is demonstrated by UOKG \cite{do-etal-2023-unsupervised}. UOKG comprises two modules, named \textit{phraseness} and \textit{informativeness}. The former, a seq2seq model trained to generate noun phrases, generate phrases while the latter, an embedding-based importance scoring function, guide this generation towards phrases that are key. Our proposed method, ERU-KG, follows this second approach.



\noindent \textbf{Generation/Extraction of keyphrases using references}. The use of references, particularly citation contexts and titles, has been explored in prior work on keyphrase extraction and generation. CiteTextRank \cite{gollapalli2014extracting} proposes a graph-based approach that incorporates citation contexts. \cite{caragea-etal-2014-citation} employ occurrences of candidates in citation contexts as a feature for supervised keyphrase extraction. \cite{garg-etal-2022-keyphrase} investigate the use of citation contexts as additional information for supervised keyphrase generation. More recently, \cite{boudin-aizawa-2024-unsupervised} proposes a framework that extracts silver-labeled keyphrases from citation contexts for domain adaptation. TG-Net \cite{chen2019guided} leverages titles to enhance input text encodings for supervised keyphrase generation. Recently, \cite{kang-shin-2024-improving} propose TPG as an unsupervised pretraining objective, where the resulting pretrained model can be viewed as an UKG model. 

Our proposed approach differs from the existing work. Specifically, our approach leverage references to learn document representations, which are used to generate keyphrases that aligned with the key concepts as recognized by the community. In contrast, existing work typically use references 1) for mining silver-labeled keyphrases or 2) as additional information to enhance the keyphrase extraction/generation process.


\noindent \textbf{Time-efficiency in keyphrase extraction and generation}. Efficient processing of large document collections is critical for the practicality of keyphrase extraction and generation models. Despite this, time-efficiency has been underdiscussed in the design of modern keyphrase extraction and generation methods. One notable exception is the work by \cite{wu2022fast}, which employs a non-autoregressive decoding strategy to significantly enhance the speed of keyphrase generation compared to autoregressive approaches. Additionally, \cite{wu2022pre} shows that prioritizing model depth over width and using deep encoders with shallow decoders has been shown to improve inference latency while maintaining accuracy. 



\section{Implementation Details}
\label{section:implementation_details}
\subsection{ERU-KG}
\label{section:implementation_details_erukg}
\noindent \textbf{Informativeness module}. We employ SPLADE as our term-importance predictor, as mentioned above. We initialized the models with DistilBERT-base\footnote{\url{https://huggingface.co/distilbert/distilbert-base-uncased}} (66M parameters) \cite{sanh2019distilbert} for ERU-KG-base and a 
BERT\_L-6\_H-512\_A-8\footnote{\url{https://huggingface.co/google/bert_uncased_L-6_H-512_A-8}} (33M parameters), which is a BERT \cite{devlin-etal-2019-bert} with 6 layers, model dimensionality of 512 and 8 attention heads, for ERU-KG-small. Models are trained with the Adam \cite{kingma2014adam} optimizer, with a learning rate of $2e^{-5}$, a warmup of 20000 steps and a batch size of 32. The models are trained for 100k steps. For FLOPS regularization, we set $\lambda_q=0.05$ and $\lambda_d=0.03$. We set the length penalty parameter, as mentioned in Eq. \ref{equation:informativeness_distribution_first_version} and \ref{equation:informativeness_distribution_second_version}, $\gamma=-0.25$.

Unless specified otherwise, the two interpolation weights $\alpha$, $\beta$ (Eq. \ref{equation:informativeness_distribution_second_version} and \ref{equation:phraseness_distribution} respectively), are both set to $0.8$. In addition, the balancing parameter $\lambda$ in Eq. \ref{equation:keyphrase_distribution}, is set to 1.5.

\noindent \textbf{Phraseness module}. We employ NLTK's \cite{bird-loper-2004-nltk} RegexpParser and extract noun phrases from document with the following grammar 

\vspace{-8pt}

\begin{equation*}
\resizebox{\columnwidth}{!}{
$(<NN.*|JJ.*>+<NN.*|CD>)|<NN.*>$
}
\end{equation*}

For finding the set of neighbor documents $\mathcal{N}(\boldsymbol{x})$ of the input text $\boldsymbol{x}$, we build BM25 retrievers using the document collection $\mathcal{D}$. In particular, $\mathcal{D}$ is the 630,749 documents from the evaluation and validation split of SciRepEval-Search\footnote{\url{https://huggingface.co/datasets/allenai/scirepeval/viewer/search}} dataset, alongside with their top 10 present keyphrases and predicted term-importances. We build our retrievers using Pyserini \cite{lin2021pyserini}. In the inference phase, we set $|\mathcal{N}(\boldsymbol{x})|=100$, unless specified otherwise.

\subsection{Keyphrase Generation/Extraction Baselines}

For TextRank and MultiPartiteRank, we use the \texttt{pke} package \cite{boudin-2016-pke}. EmbedRank is implemented following the description in \cite{bennani-smires-etal-2018-simple}, with the exception that we employ the same noun phrase extractor described in \ref{section:implementation_details_erukg}. For EmbedRank, we employ both Sent2Vec (sent2vec\_wiki\_unigrams\footnote{\url{https://github.com/epfml/sent2vec}}) \cite{pagliardini-etal-2018-unsupervised}, as in the original paper, and SBERT (all-MiniLM-L12-v2\footnote{\url{https://huggingface.co/sentence-transformers/all-MiniLM-L12-v2}}) \cite{reimers-gurevych-2019-sentence}. For PromptRank \cite{kong-etal-2023-promptrank}, we adopt the official implementation\footnote{\url{https://github.com/NKU-HLT/PromptRank}}.

For AutoKeyGen, UOKG, and CopyRNN, we use the implementations and checkpoints provided by the authors of \cite{do-etal-2023-unsupervised} \footnote{\url{https://github.com/ForwardDataLab/UOKG/issues/1}}. Finally, for TPG\footnote{\url{https://github.com/kangnlp/low-resource-kpgen-through-TPG}} \cite{kang-shin-2024-improving} and PromptKP\footnote{\url{https://github.com/m1594730237/FastAndConstrainedKeyphrase}} \cite{wu2022fast}, we utilize the official implementation.

\subsection{RM3 and DocT5Query}
For DocT5Query, we utilized the pre-generated queries provided for the datasets within the BEIR benchmark. For RM3, we leveraged Pyserini's \cite{lin2021pyserini} implementation\footnote{\url{https://github.com/castorini/pyserini}} and utilize the default hyperparameters.

\subsection{Computing Infrastructure}
\label{section:computing_infrastructure}
We run all our experiments on a server with two AMD EPYC 7302 3GHz CPUs, three NVIDIA Ampere A40 GPUs (300W, 48GB VRAM each), and 256 GB of RAM.

\section{Case Study}
\label{section:case_studies}

To gain further insights into ERU-KG's effectiveness, we display the keyphrases generated by ERU-KG and the baselines on two types of text, namely \textit{document} and \textit{query}, in Table \ref{table:document_case_study}
 and Table \ref{table:query_case_study}, respectively. For \textit{document}, we use the same example document as in Figure \ref{fig:ReferencesExample}. For \text{query}, we provide two examples, a long multi-aspected query from DORIS-MAE dataset and a short query from SCIDOCS. 

 \noindent \textbf{Document}. Upon initial examination, there appear to be no significant differences in the predicted present keyphrases across methods, as they all reflect concepts used in reference to the given document. However, considering absent keyphrases, ERU-KG produces keyphrases that are more relevant. Specifically, ERU-KG is able to predict ``sea ice classification'', ``sea ice concentration'' and ``sea ice detection'', which are not only used later in the main body of the given paper, but also used in a citation context (``sea ice classification'' is used in the second citation context in Table \ref{table:document_case_study})
 
 

\noindent \textbf{Query}. It can be seen that keyphrases generated by ERU-KG might be more beneficial as additional information. In the first example, ERU-KG is the only model that can produce the name of alternative GAN techniques (e.g. ``ac gan'', ``am gan'', ``net gan'' and ``conditional gan''). Moreover, the introduction of phrases such as ``image generation'' and ``synthetic data'' is also suitable for the objective of the user. 

\begin{table}[h]
\centering

\resizebox{\columnwidth}{!}{
    \begin{tabular}{|l|cc|cc|}
    \hline
    \multirow{2}{*}{}               & \multicolumn{2}{c|}{\begin{tabular}[c]{@{}c@{}}Present keyphrase \\ generation\end{tabular}} & \multicolumn{2}{c|}{\begin{tabular}[c]{@{}c@{}}Absent keyphrase \\ generation\end{tabular}} \\ \cline{2-5} 
                                    & F@5                                           & F@10                                         & R@5                                          & R@10                                         \\ \hline
    $\alpha, \beta = 0$   & 13.9                                          & 12.1                                         & 5.1                                          & 7.1                                          \\
    $\alpha, \beta = 0.2$ & 17.7                                          & 18.2                                         & 5.6                                          & 7.6                                          \\
    $\alpha, \beta = 0.4$ & 20.8                                          & 19.9                                         & 5.8                                          & 7.9                                          \\
    $\alpha, \beta = 0.6$ & 22                                            & 19.9                                         & 5.9                                          & 8.1                                          \\
    $\alpha, \beta = 0.8$ & 22                                            & 19.4                                         & 6                                            & 8.1                                          \\
    $\alpha, \beta = 1$   & 21.5                                          & 18.8                                         & 0                                            & 0                                            \\ \hline
    \end{tabular}
}

\caption{Sensitivity analysis of interpolation hyperparameters $\alpha$ and $\beta$. Experiments are conducted on the KP20K dataset, using ERU-KG-base.}

\label{table:sensitivity_analysis}
\end{table}

In the second example, ERU-KG is the only KG model that manages to generate ``brain computer interface'' - the full-form version of ``BCI''. In addition, other absent phrases predicted by ERU-KG, e.g. ``domain adaptation'', ``meta learning'', are also highly relevant. On the other hand, it can be seen that absent keyphrases generated by other KG methods do not offer as much valuable additional information. In particular, AutoKeyGen and UOKG produces absent keyphrases that are oftenly reorderings of present terms, while CopyRNN introduces irrelevant keyphrases, such as ``world wide web''.

\section{Additional Ablation Studies}

\subsection{Sensitivity Analysis of Interpolation Hyperparameters}
\label{section:sensitivity_analysis_interpolation}

To better understand the impact of the hyperparameters $\alpha$ and $\beta$ on keyphrase generation quality, we conduct experiments on the KP20K dataset and using the ERU-KG-base model. In our experiments, we set the two hyperparameters to the same value, as both controls the influence of related documents. The results for both present and absent keyphrase generation are displayed in Table \ref{table:sensitivity_analysis}.

Firstly, when $\alpha, \beta=0$ (i.e. only information from related documents are used), the performance is inferior across all metrics. Secondly, at the opposite extreme where $\alpha, \beta = 1$ (i.e. only information from the input document is used), there exists two limitations: 1) no absent keyphrase are predicted and 2) keyphrase extraction performance drops slightly, indicating that related documents are beneficial for this task. Finally, across the intermediate range of $\alpha, \beta = 0.2$ to $0.8$, we observe an upward trend in performance. Based on the above discussions, we conclude that optimal performance is achieved when both sources are utilized, with the given document maintains greater influence.

\section{Algorithm Descriptions of ERU-KG}
We provide an algorithm description of the inference process of ERU-KG in Algorithm \ref{algorithm:eru_kg_inference}.

\begin{algorithm*}
  \caption{ERU-KG inference}
  \label{algorithm:eru_kg_inference}

  \Input{Document $\boldsymbol{x}$, number of output keyphrases $k$}
  \Output{Sets of present and absent keyphrases $\boldsymbol{Y}_{\boldsymbol{x}}^{\text{present}}$ and $\boldsymbol{Y}_{\boldsymbol{x}}^{\text{absent}}$, each containing $k$ keyphrases}

  \BlankLine
\hrule
\BlankLine

\nllabel{}\centerline{\textbf{Phraseness module}}

\BlankLine
\hrule
\BlankLine

$\mathcal{N}(\boldsymbol{x}), \{\tilde{\boldsymbol{s}}_{\boldsymbol{x}, \boldsymbol{x}'} \mid \boldsymbol{x}' \in \mathcal{N}(\boldsymbol{x})\} \gets \text{BM25Retrieve}(\text{query} = \boldsymbol{x}, \text{numdocs}=100)$ \Comment*[r]{Retrieve similar documents and the similarity scores}
  
  $\boldsymbol{C}_{\boldsymbol{x}} \gets \text{NounphraseExtract}(\boldsymbol{x})$


 $\boldsymbol{C}_{\mathcal{N}(\boldsymbol{x})} \gets \{\}$
  
  \ForEach{$\boldsymbol{x}' \in \mathcal{N}(\boldsymbol{x})$}{

  $\tilde{\boldsymbol{C}}_{\boldsymbol{x}'} \gets \text{GetPrecomputedCandidate}(\boldsymbol{x}')$

  $\boldsymbol{C}_{\mathcal{N}(\boldsymbol{x})} \gets 
 \boldsymbol{C}_{\mathcal{N}(\boldsymbol{x})} \cup \tilde{\boldsymbol{C}}_{\boldsymbol{x}'}$


  }








$\hat{\boldsymbol{C}}_{\boldsymbol{x}} \gets \boldsymbol{C}_{\boldsymbol{x}} \ \cup \text{Top}_{100}(\boldsymbol{C}_{\mathcal{N}(\boldsymbol{x})}, P_{\text{pn}})$
  

    \BlankLine
\hrule
\BlankLine

\nllabel{}\centerline{\textbf{Informativeness module}}

\BlankLine
\hrule
\BlankLine

  $w^{\boldsymbol{x}} =\{w^{\boldsymbol{x}}_j\}_{j \in V} \gets \text{SPLADE}(\boldsymbol{x})$ \Comment*[r]{Term importances given $\boldsymbol{x}$. $V$ denotes BERT's vocabulary }

    
    \ForEach{$\boldsymbol{x}' \in \mathcal{N}(\boldsymbol{x})$}{

    $w^{\boldsymbol{x}'}=\{w^{\boldsymbol{x}'}_j\}_{j \in V} \gets \text{SPLADE}(\boldsymbol{x}')$ \Comment*[r]{Precomputed}

    }

    \ForEach{$j \in V$}{

    $\hat{w}^{\boldsymbol{x}}_{j} \gets \alpha \ w^{\boldsymbol{x}}_{j} + (1-\alpha) \sum_{\boldsymbol{x}' \in \mathcal{N}(\boldsymbol{x})}\tilde{s}_{\boldsymbol{x},\boldsymbol{x}'} \ w^{\boldsymbol{x}'}_{j}$
    }

  \ForEach{$\boldsymbol{c} \in \hat{\boldsymbol{C}_{\boldsymbol{x}}}$}{

    $\hat{f}(\boldsymbol{c}, \boldsymbol{x}) \gets \frac{1}{|\boldsymbol{c}| - \gamma} \sum_{i=1}^{|\boldsymbol{c}|} \hat{w}_{\boldsymbol{x}}(c_i)$
  }

  \BlankLine
\hrule
\BlankLine

\nllabel{}\centerline{\textbf{Combining phraseness and informativeness}}

\BlankLine
\hrule
\BlankLine

  \ForEach{$\boldsymbol{c} \in \hat{\boldsymbol{C}_{\boldsymbol{x}}}$}{

    $P_{\text{in}}(\boldsymbol{c}|\boldsymbol{x}) \gets \hat{f}(\boldsymbol{c}, \boldsymbol{x}) / \sum_{\boldsymbol{c}' \in \hat{\boldsymbol{C}}_{\boldsymbol{x}}} \hat{f}(\boldsymbol{c}', \boldsymbol{x})$ \Comment*[r]{Since the final score is only used for ranking, we skip this normalization step in practice and directly set $P_{\text{in}}(\boldsymbol{c}|\boldsymbol{x}) \gets f^{\text{in}}_{\boldsymbol{x}}(\boldsymbol{c})$}

    \;
    
    $P_{\text{kp}}(\boldsymbol{c}|\boldsymbol{x}) \gets P_{\text{pn}}(\boldsymbol{c}|\boldsymbol{x})^{\lambda} \times P_{\text{in}}(\boldsymbol{c}|\boldsymbol{x})$ \Comment*[r]{Keyphrase distribution given $\boldsymbol{x}$. $P_{\text{kp}}(\boldsymbol{c}|\boldsymbol{x})$ is also not normalized since we only use it for ranking}

    \;

  $\boldsymbol{s}_{\boldsymbol{x}}(\boldsymbol{c}) \gets \omega_{\boldsymbol{x}}(\boldsymbol{c}) P_{\text{kp}}(\boldsymbol{c}|\boldsymbol{x})$ \Comment*[r]{Apply position penalty}
  }

  $\boldsymbol{Y} \gets \texttt{sorted}(\hat{\boldsymbol{C}_{\boldsymbol{x}}}, \texttt{sortby} = \boldsymbol{s}_{\boldsymbol{x}}(\boldsymbol{c}), \texttt{descending=True})$

  $\boldsymbol{Y}_{\boldsymbol{x}}^{\text{present}} = \{\boldsymbol{y} \in \boldsymbol{Y} \mid \boldsymbol{y} \in \boldsymbol{x}\}[:k]$
  
  $\boldsymbol{Y}_{\boldsymbol{x}}^{\text{absent}} = \{\boldsymbol{y} \in \boldsymbol{Y} \mid \boldsymbol{y} \not \in \boldsymbol{x}\}[:k]$

\end{algorithm*}

\begin{table*}[]
    \centering
    \includegraphics[width=\textwidth]{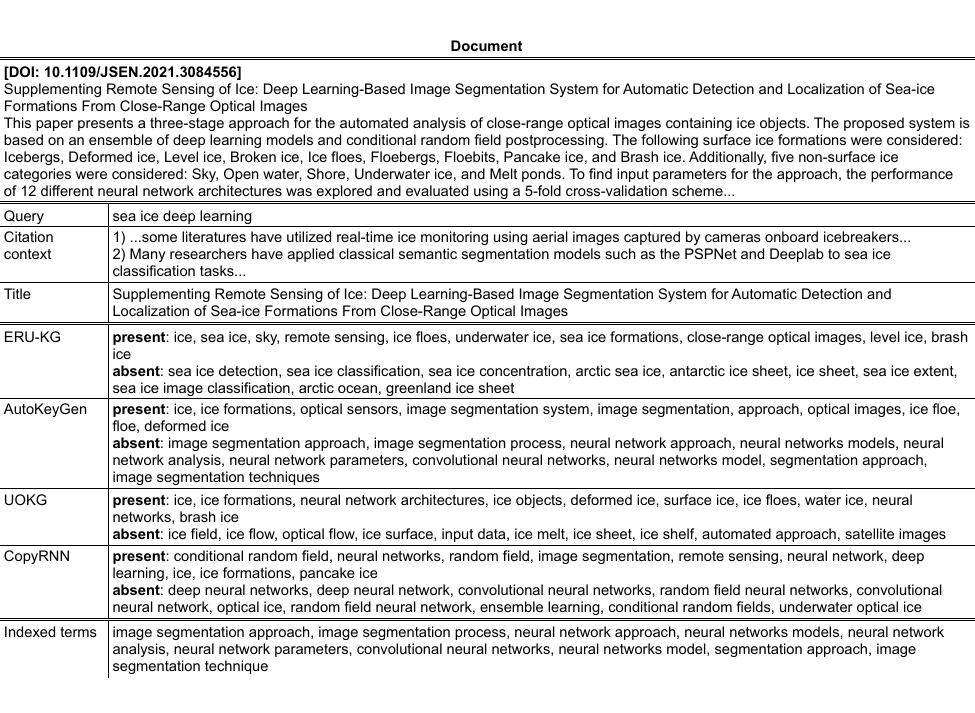}
    \caption{Generated keyphrases for an example document, by our proposed model and the baselines. We illustrate the
top 10 present and absent keyphrases. In addition, we provide the paper's indexed terms, as well as references of each type (i.e. query, citation context and title) that mentions the given paper.}
    \label{table:document_case_study}
\end{table*}

\begin{table*}[]
    \centering
    \includegraphics[width=\textwidth]{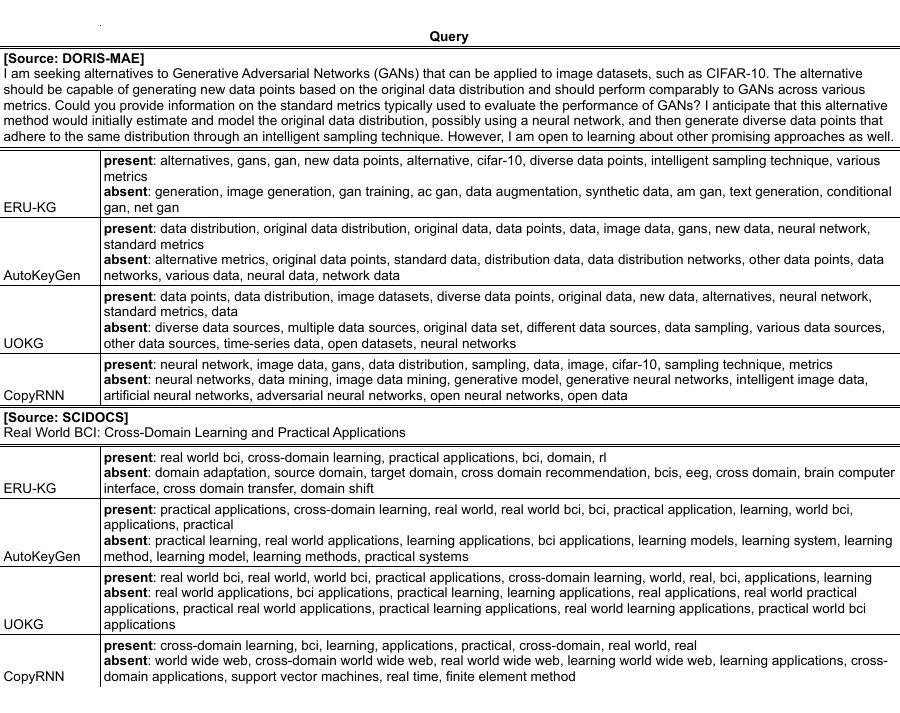}
    \caption{Generated keyphrases for two example queries, by our proposed model and the baselines. We illustrate the
top 10 present and absent keyphrases.}
    \label{table:query_case_study}
\end{table*}

\end{document}